# AttnRegDeepLab: A Two-Stage Decoupled Framework for Interpretable Embryo Fragmentation Grading

Ming-Jhe Lee
Department of Electrical Engineering
National Taiwan Ocean University
Keelung 202301, Taiwan
jerry80098009@gmail.com

Chang-Hong Wu
Department of Electrical Engineering
National Taiwan Ocean University
Keelung 202301, Taiwan
s0979979782@gmail.com

Jung-Hua Wang*
AI Research Center
National Taiwan Ocean University
Keelung 202301
jhwang@email.ntou.edu.tw

Ming-Jer Chen
Department of Obstetrics, Gynecology and Woman's Health, TVGH, 407219
Taichung, Taiwan
mingjerchen@gmail.com

Yu-Chiao Yi
Department of Obstetrics, Gynecology and Woman's Health, TVGH, 407219
Taichung, Taiwan
yuchiaoyi@gmail.com

Tsung-Hsien Lee
Department of Obstetrics and Gynecology, CSMU Hospital,
Taichung 40201, Taiwan
jackth.lee@gmail.com

***Abstract***—Assessing embryo fragmentation is crucial for predicting IVF success, yet manual grading is prone to subjectivity, and existing AI models struggle with clinical interpretability and segmentation errors. We propose AttnRegDeepLab, a Multi-Task Learning (MTL) framework designed to solve these challenges. The model enhances a DeepLabV3+ decoder with Attention Gates to filter out cytoplasmic noise and retain sharp contour details. It also introduces a Multi-Scale Regression Head with Feature Injection, guiding the segmentation process with global grading priors to eliminate systematic area estimation errors. Based on a two-stage decoupled training strategy and a range-based loss for weakly labeled data, our method resolves MTL gradient conflicts. AttnRegDeepLab yields high grading precision and excellent segmentation quality (Dice coefficient = 0.729), avoiding the trade-off between contour integrity and grading accuracy seen under standard joint optimization. This provides a reliable, clinically interpretable tool balancing visual and quantitative accuracy.

***Keywords***: Embryo Grading, Multi-Task Learning, Explainable AI, Gradient Conflict, Semantic Segmentation, Decoupled Training.

## I. Introduction

### A. Background

Infertility is a critical global public health issue, with assisted reproductive technology (ART)—specifically in vitro fertilization (IVF)—serving as the primary clinical intervention. IVF success depends heavily on the embryo transfer outcome, which is largely dictated by embryo morphological quality [1]. Among key morphological indicators, embryo fragmentation—the presence of anucleate cytoplasmic fragments during early cleavage—is a major negative prognostic factor that impairs developmental potential and reduces implantation rates [2]. Consequently, precise quantitative and qualitative assessment of fragmentation is vital for optimizing embryo selection and improving ART efficacy.

While clinical practice standardly utilizes the Gardner grading system (or similar frameworks) to categorize fragmentation into discrete tiers—Grade A (<10%), Grade B (10–25%), Grade C (25–50%), and Grade D (>50%)—this manual evaluation is inherently subjective. Significant inter-observer variability exists even among experienced embryologists [3,4]. This subjectivity introduces risks of misclassification for viable embryos, ultimately compromising clinical decision-making and patient outcomes.

### B. Motivation and Technical Challenges

To mitigate the limitations of manual assessment, Computer-Aided Diagnosis (CAD) systems have been introduced. However, existing automated solutions face significant technical bottlenecks in clinical deployment. First, traditional texture analysis methods (e.g., GLCM) lack the feature discriminability required to process complex embryonic cytoplasmic textures, often failing to distinguish pathological fragments from normal blastomeres. Second, while deep learning-based regression models can predict fragmentation grades, they function as "black boxes." These models fail to provide visual evidence of fragment localization, falling short of the stringent explainability standards required in medical AI.

To enhance interpretability, an ideal system must simultaneously output a quantitative grade and a precise segmentation mask. Yet, multi-task optimization reveals an inherent conflict: segmentation prioritizes pixel-level boundary fidelity, whereas grading relies on global area estimation. In shared-backbone architectures, this divergence frequently induces negative transfer [5,6], where the optimization of one task compromises the performance of the other. To overcome this pervasive multi-task learning challenge in medical imaging, this paper proposes an integrated framework designed for precise, explainable embryo fragmentation analysis.

### C. Contributions

The primary contributions of this work are threefold: (1) we introduce an integrated framework that leverages both global and local features by coupling an Attention-Guided DeepLabV3+ backbone with a Multi-Scale Regression Head. A novel feature-injection mechanism feeds global grading priors back into the segmentation decoder, systematically correcting area estimation biases; (2) To mitigate gradient interference and joint-optimization instability, we develop a decoupled training strategy consisting of a pretrain–freeze–finetune pipeline that stabilizes multi-task learning; (3) To address the scarcity of pixel-level annotations, we implement a range-based loss function that converts discrete clinical grades into continuous interval constraints. This allows extensive weakly labeled datasets to be effectively exploited within a semi-supervised paradigm.

## II. RELATED WORK

This section reviews related embryo grading techniques and, utilizing preliminary experimental data from this study, analyzes the limitations of traditional methods and existing deep learning strategies. This analysis establishes the necessity of the proposed multi-task learning and decoupled training framework.

### *A. Limitations of Hand-crafted Features*

Historically, texture analysis was the primary method for quantifying embryo quality. Since fragmentation manifests as textural anomalies disrupting cytoplasmic uniformity, verifying the efficacy of classic descriptors, specifically the Gray-Level Co-occurrence Matrix (GLCM), is a necessary prerequisite to justify the transition to deep learning.

#### 1) Data Selection and Clinical Relevance

To ensure clinical relevance, preliminary experiments focused on the early cleavage stage, utilizing embryo images at the T2 (2-cell) and T4 (4-cell) stages. This selection is justified by two primary factors: first, literature establishes that the degree of cleavage-stage fragmentation (Days 2–3) is a critical biomarker for predicting blastocyst development potential [7,8]. Second, compared to compacted morulae, T2 and T4 embryos exhibit distinct cell boundaries and fragment structures, providing an ideal baseline to evaluate an algorithm's capacity to differentiate between cytoplasmic noise and true fragmentation.

#### 2) Feature Space Overlap (GLCM Analysis)

Using the T2/T4 dataset, Gray-Level Co-occurrence Matrix (GLCM) texture features were extracted from the embryonic regions of interest (ROIs) and visualized via t-SNE. In Fig. 1, even polar grades (Grade A vs. Grade D) exhibit severe feature-space overlap. This demonstrates that GLCM features lack the discriminability required to resolve common background artifacts in early cleavage images—such as cytoplasmic granules and vacuoles—thereby failing to establish distinct decision boundaries.

#### 3) Failure of Non-linear Mapping

To address GLCM limitations, evaluations were extended to Local Binary Patterns (LBP) [9] for enhanced local detail sensitivity, coupled with a Multi-Layer Perceptron (MLP) for non-linear regression. While the MLP identified extreme fragmentation (Grade D), it failed to differentiate subtle variations among intermediate tiers (Grades A, B, and C), as shown in Fig. 2. This confirms that hand-crafted features are fundamentally inadequate for capturing nuanced embryonic morphological changes, underscoring the necessity of deep-learning-driven automated feature representation.

### *B. Baselines*

Next, we evaluate two common strategies and identify their respective flaws.

#### 1) Pure Regression Models and the "Black-Box" Flaw

Standard Convolutional Neural Networks (CNNs), such as InceptionV3, were trained as regression models to directly predict fragmentation ratios. While effective convergence and alignment with ground-truth trends (Fig. 3) confirmed that CNNs can successfully extract grading features, this approach suffers from a fundamental clinical limitation: the "black-box" nature of deep learning. Because the model outputs only a numerical score without visual evidence of fragment localization, it fails to meet the stringent explainability standards required in medical AI. Furthermore, feature visualization revealed that the lack of pixel-level supervision caused the network to frequently attend to non-fragment background noise, preventing clinicians from visually verifying the basis of the grading.

#### 2) Pure Segmentation Models and Accumulated Errors in Indirect Grading

To address the lack of interpretability, an alternative paradigm employs semantic segmentation (e.g., DeepLabV3+) to isolate fragment regions and calculate the fragmentation ratio via indirect grading. However, experiments revealed that the Mean Absolute Error (MAE) of this approach is highly sensitive to uncertainties in boundary delineation. Because embryo fragment boundaries are typically blurred and irregular, minor pixel-level segmentation errors at the contours accumulate into significant grading deviations during global area integration. This demonstrates that optimizing the segmentation loss ($L_{seg}$) alone is insufficient to guarantee numerical grading precision.

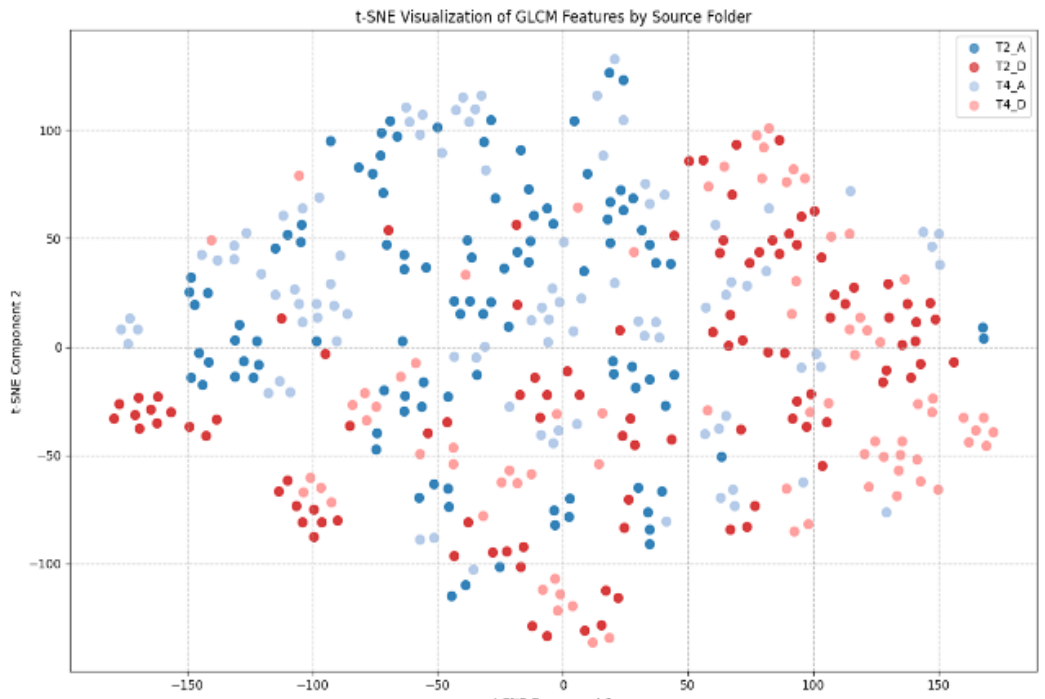

Fig. 1. t-SNE visualization of GLCM texture features.

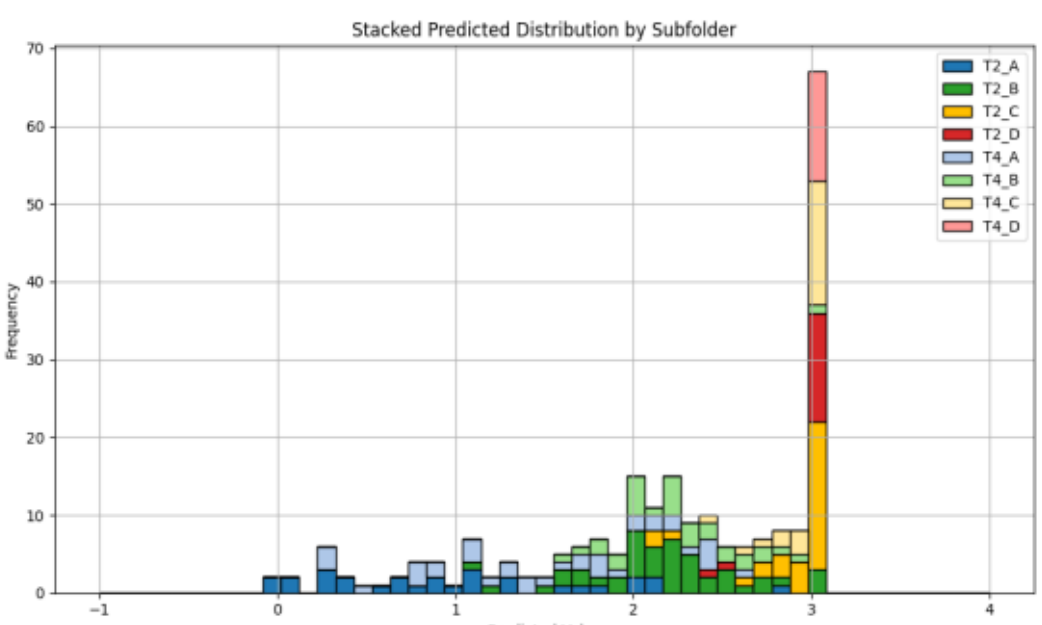

Fig. 2. Prediction histogram of LBP-based MLP model.

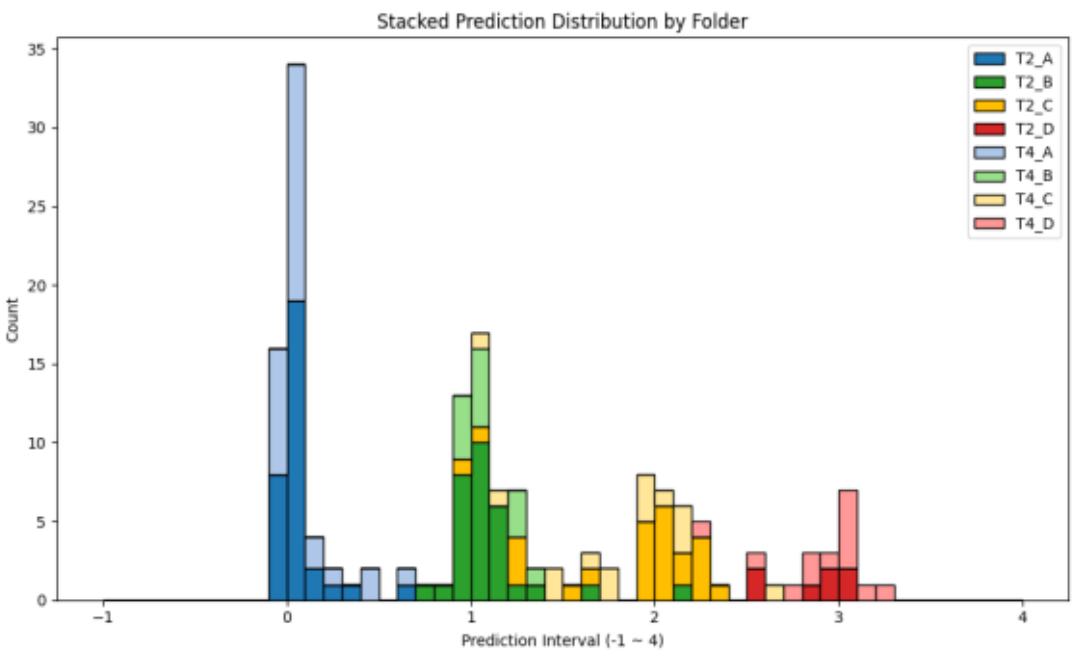

Fig. 3. Distribution of predicted grades by the pure regression baseline.

### C. Multi-Task Learning and Gradient Conflict

The above analysis indicates that a practical system should at least combine the following two elements: simultaneously outputting numerical grading and pixel-level segmentation masks. This observation aligns with the emerging paradigm of Multi-Task Learning (MTL) in medical imaging [10].

However, past literature [5, 6] indicated that when there is a *competitive* relationship between tasks (e.g., the segmentation task pursues high-frequency boundary details, while the regression task pursues low-frequency global abstractions), the backbone faces *gradient conflict*. Namely, the segmentation gradient $L_{seg}$ tends to optimize for spatial details, whereas the regression gradient $L_{reg}$ tends to optimize for global abstraction. This conflict can easily induce Negative Transfer in the shared backbone, where the feature representation for segmentation is compromised to optimize regression.

### D. Issue of Discrete Grades

Clinically, grades A, B, C, and D represent distinct intervals within a continuous proportional scale (0–100%) [11]. Treating the grading task as a standard one-hot encoding problem would discard critical magnitude and ordinal information. To address this, we reformulate the grading task as a continuous ratio regression problem and utilize discrete labels to establish explicit interval constraints. This approach is analogous to the granular evaluation strategy (i.e., first segmentation, then classification) proposed by [12]. Our approach aligns more closely with clinical reality and ensures stable model convergence.

## III. Methodologies

This section details AttnRegDeepLab, which is a dual-branch architecture combining an attention-guided DeepLabV3+ with a multi-scale regression head, characterized by a feature injection mechanism alongside a two-stage decoupled training strategy.

### A. System Architecture Design

In Fig. 4, the framework integrates a shared backbone with two task-specific branches: a regression module for global quantification and an attention-guided segmentation module for local boundary delineation.

*1) Shared Backbone and Multi-Scale Features.* We adopt ResNet-50 as the shared feature extractor. To accommodate the spatial resolution requirements of semantic segmentation, we modified the standard architecture by removing the downsampling stride in Layer 3 and Layer 4. We used dilated convolutions, with rate values of 2 and 4 at layer 3 and layer 4, respectively, to maintain the output stride of the final feature map ($F_4$) at 8. This modification allows the backbone to preserve high-level semantic abstraction without losing essential spatial details. The model extracts features from multiple levels for downstream tasks, specifically utilizing the low-level $F_1$ features (Layer 1, 1/4 scale, 256 channels) for boundary recovery, and the high-level $F_3$ (Layer 3, 1024 channels) and $F_4$ (Layer 4, 2048 channels) features as the core inputs for segmentation context and regression analysis.

*2) Regression Module:* To map input images to a continuous grading scale $\hat{y}_{reg} \in [0, 1]$, we designed a Multi-Scale Regression Head. Features from layers $F_3$ and $F_4$ are fused and processed via Global Average Pooling (GAP), followed by MLP. We extracted a latent feature vector $V_{reg} \in \mathbb{R}^{2048}$ from the penultimate layer of the MLP. This vector serves as a compact representation of global fragmentation degree, which is injected into the segmentation branch to guide pixel-level predictions.

*3) Attention-Guided Segmentation Module:* To overcome the over-smoothing inherent in standard DeepLabV3+ and correct estimation errors, we redesigned the decoder into a three-stage refinement pipeline:

**Multi-Scale Context Extraction:** First, the backbone features ($F4$) are processed through Atrous Spatial Pyramid Pooling (ASPP). This captures multi-scale contextual information, yielding a high-level semantic map $F_{ASPP}$.

**Attention-Guided Boundary Refinement:** To recover fine spatial details lost during down-sampling without introducing cytoplasmic noise, we replaced the standard skip connection with an Attention Gate (AG). Let $F_1 \in R^{C_l \times H \times W}$ denote the shallow features from the backbone and $g = \mathrm{Up}(F_{ASPP})$ be the upsampled gating signal. The attention coefficient map $\alpha$ is formulated as

$$\alpha = \sigma\left(\psi\left(\mathrm{ReLU}\left(W_x(F_1) + W_g(g) + b\right)\right)\right) \quad (1)$$

The refined shallow features are then obtained by $\widehat{F_1} = \alpha \odot F_1$, which effectively filters out background texture noise before concatenation.

**Regression-Guided Feature Injection:** Finally, to align the pixel-level segmentation with the global severity grade,

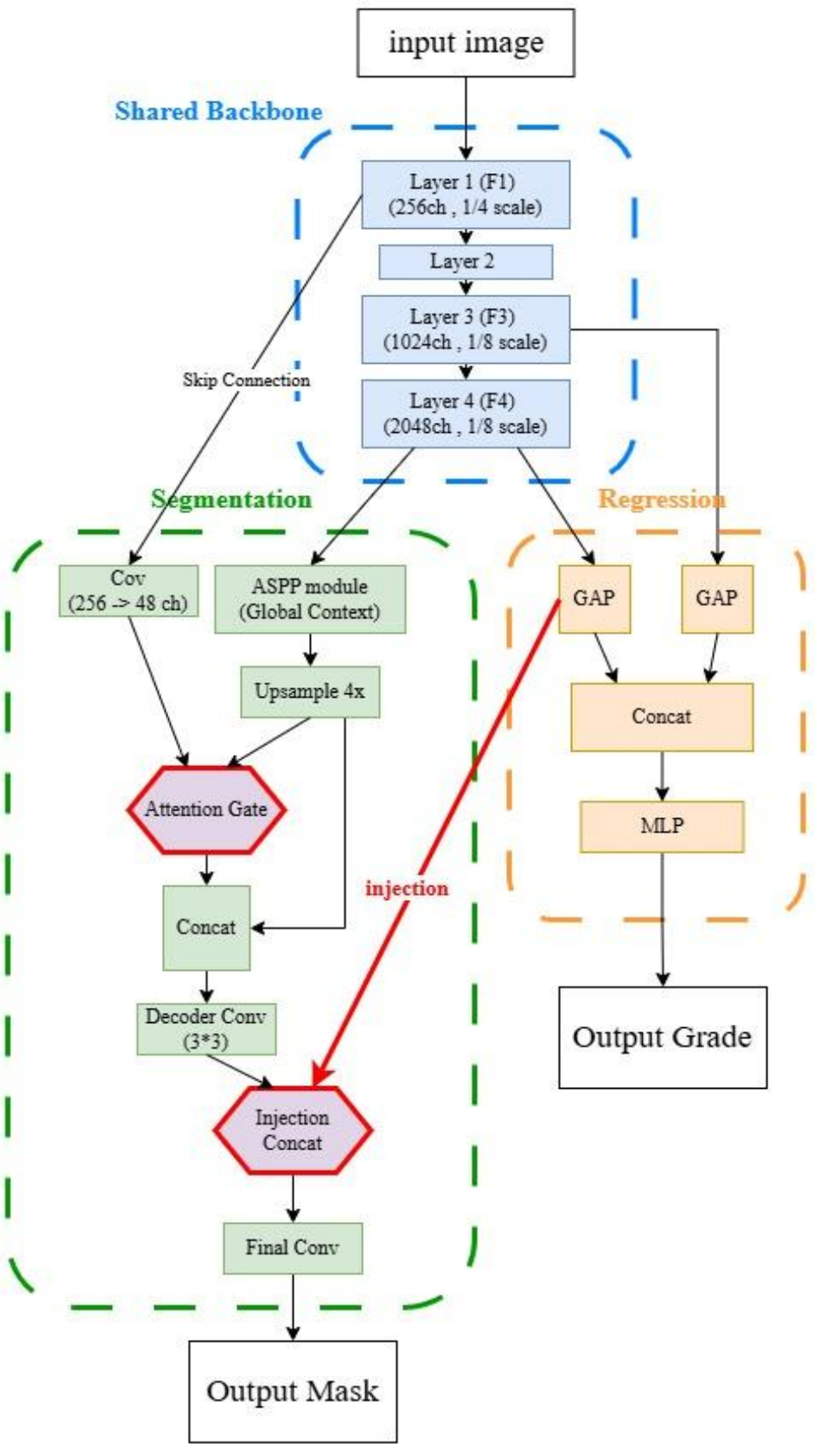

Fig. 4. Schematic overview of AttnRegDeepLab, comprising a shared ResNet-50 backbone (blue), an attention-guided segmentation branch (green), and a multi-scale regression branch (orange). The red arrow indicates the U-Net skip connection, while the 'Injection' module represents the fusion bridge.

a Feature Injection mechanism is applied at the end of the decoder. The global regression vector $V_{reg}$ is spatially broadcast and fused with the decoder features $F_{dec}$:

$$F_{\text{inj}} = \text{Broadcast}(V_{\text{reg}}) \in R^{C_{\text{reg}} \times H \times W} \quad (2)$$

$$F_{\text{final}} = \text{Concat}(F_{\text{dec}}, F_{\text{inj}}) \quad (3)$$

By conditioning the final convolution on $F_{final}$, the model incorporates global grading priors to rectify local segmentation ambiguities.

## B. Hybrid Loss Function

To coordinate the pixel-level and image-level tasks under semi-supervised settings, we minimize a composite objective:

$$L_{\text{total}} = \alpha L_{\text{seg}} + \beta L_{\text{reg}} + \gamma L_{\text{cons}} \quad (4)$$

where $L_{seg}$ is defined as the weighted sum of Binary Cross Entropy, Dice Loss [13], and Focal Loss. It is aimed at addressing class imbalance between sparse fragments and the background. For $L_{reg}$, we adopt a dual-strategy. For samples possessing complete pixel-level supervision (referred to as paired data, $D_{paired}$), we employ a precise regression loss (denoted as $L_{pre}$) and calculate the *L1* error against the ground truth ratio derived from pixel-level annotations, whereas for samples limited to clinical grades only ($D_{weak}$), we will use a Range Loss ($L_{range}$), making it consistent with emerging weak supervision paradigms in medical imaging [14]. While mapping each discrete grade $G$ to a continuous interval [$y_{min}$, $y_{max}$], the loss penalizes predictions $\hat{y}$ only when they violate interval bounds:

$$L_{\text{range}}(\hat{y}, G) = \text{ReLU}(y_{\min} - \hat{y}) + \text{ReLU}(\hat{y} - y_{\max}) \quad (5)$$

Eq.(5) enables the utilization of large-scale clinical records without enforcing arbitrary point-labels. Finally, for $L_{cons}$, we minimize the discrepancy between the regression prediction and the area ratio derived from the segmentation mask, ensuring that geometrical and numerical outputs are mutually reinforcing.

## C. Two-Stage Decoupled Training Strategy

To mitigate the negative transfer, we propose a sequentially decoupled training strategy. In Phase 1 (Visual Expert Pre-training), the objective is to train the backbone network's ability to extract spatial features and optimize boundaries using the Attention mechanism. In this phase, we enable only $L_{seg}$ and $L_{cons}$ ($\beta$=0) and initialize the model with ImageNet weights. This results in a feature extractor with optimal visual fidelity, establishing a baseline for instance segmentation. In Phase 2 (Regression-Guided Finetuning), we address the gradient conflict. We freeze the parameters of the shared backbone $\theta_{\text{bckbn}}$ learned in Phase 1, treating it as a fixed feature extractor. Optimization is restricted to the regression head $\theta_{\text{reg}}$ and the feature injection layers..

$$\theta_{\text{bckbn}}^{(\text{t}+1)} = \theta_{\text{bckbn}}^{(\text{t})} \quad (\text{Frozen}). \quad (6)$$

Eq.(6) prevents the regression gradients, which prioritize global quantity, from disrupting the high-frequency spatial features essential for segmentation boundaries, effectively isolating the negative transfer.

# IV. EXPERIMENTAL RESULTS

To verify the effectiveness of AttnRegDeepLab, rigorous ablation studies were conducted. We first describe the experimental setup, followed by a step-by-step analysis of component removal to evaluate the specific impact of the Feature Injection mechanism and the Two-Stage Decoupled strategy on model performance.

## A. Experimental Setup

*1) Data Source and Partitioning:* This study focused on the critical cleavage stage (T2 and T4). Dataset was split into two subsets. First, Paired Data ($D_{paired}$) with 318 samples annotated with both pixel-level fragmentation masks and clinical grades. This subset supports fully supervised training for $L_{seg}$ and $L_{pre}$. Secondly, Grade-Only Data ($D_{weak}$) of 1549 samples annotated with clinical grades without segmentation masks. This subset is utilized exclusively for weakly supervised regression via $L_{range}$.

*2) Preprocessing:* To prevent geometric distortion, all input images were preprocessed using an aspect-ratio preserving resizing with zero-padding, standardizing the resolution to 299 × 299 pixels.

*3) Implementation Details:* Using PyTorch and the AdamW optimizer on NVIDIA RTX 3060, our framework relies on a differential learning rate schedule to stabilize the decoupled training. In the initial unfrozen phase (full MTL), we applied a learning rate of $1\times10^{-4}$ to extract global features. During the subsequent decoupled fine-tuning phase, we froze the backbone and lowered the rate to $1\times10^{-5}$ to safeguard the learned feature space.

*4) Evaluation Metrics:* Performance was assessed using (1) Dice Similarity Coefficient (DSC) to measure the spatial overlap between the predicted segmentation mask and the expert annotation; (2) Mean Absolute Error (MAE) to evaluate the accuracy of the regression head in predicting the fragmentation ratio against the ground truth.

## B. Preliminary Study

To confirm the benefits of integrating Attention Gates into the U-Net skip connections, we compared our architecture against the Standard DeepLabV3+ baseline in a single-task setting. Experimental results (Table 1, Rows 1&2) indicate that while the Standard DeepLabV3+ benefits from the ASPP module for global semantics, its segmentation accuracy stagnates at Dice=0.717. Qualitative analysis reveals a tendency towards over-smoothing and misclassification of cytoplasmic granules near blurred fragment boundaries. In contrast, by incorporating attention-guided shallow features ($F_1$), our method increased the segmentation accuracy to 0.729 (+1.2%). This confirms that filtering high-frequency noise via Attention Gates is critical for improving boundary discriminability in embryo fragmentation, and justifies that AttnRegDeepLab is a feasible baseline for subsequent multi-task experiments.

## C. Ablation Study and Core Analysis

A series of ablation experiments were conducted to systematically quantify the gradient trade-off in multi-task learning.

*1) Experiment settings:* To systematically isolate the contribution of each component, we defined five

configurations alongside the Standard DeepLabV3+ Baseline (see Table 1):

**Exp A (Visual Baseline/S1):** The segmentation branch trained independently with optimization restricted to segmentation objectives ($L_{seg}$ + $L_{cons}$,). This setup aims to establish the upper bound for segmentation accuracy (S1) without interference from regression gradients.

**Exp B (Regression Baselines):** Consists of pure regression models to benchmark quantification accuracy in the absence of spatial supervision. To validate the efficacy of Range Loss, we stratify this into Exp B1 (Paired Only), trained exclusively on the $D_{paired}$ subset, and Exp B2 (Semi-supervised), trained on the full dataset ($D_{paired}$ + $D_{weak}$). Comparing B1 and B2 allows us to quantify the explicit benefit of incorporating weakly labeled data before integrating it into the full multi-task framework.

**Exp C (Full MTL):** Implements the complete AttnRegDeepLab with feature injection under joint optimization by minimizing $L_{total}$. This experiment is designed to analyze the shape-area tradeoff under joint optimization and assess the impact of gradient conflict.

**Exp D (Decoupled Strategy):** Adopts the Pretrain-Freeze-Finetune strategy. By optimizing $L_{seg}$ in Phase 1 and $L_{reg}$ in Phase 2 (with a frozen backbone), this experiment verifies the hypothesis of positive transfer, testing whether fixed spatial features can support accurate regression without catastrophic forgetting.

**Exp E (No-Inject Control):** Serves as a strict ablation control for Exp C where the feature injection mechanism is disconnected, while maintaining the same joint loss function ($L_{total}$). Comparing Exp C and Exp E allows us to isolate the specific contribution of the physical feature bridge in coordinating the two tasks.The quantitative results of these experiments are summarized in Table 1.

Table 1: Quantitative Results of Ablation Study

| Exp ID | Architecture & Strategy | Strategy | Loss Setting | Dice Score (F1) | MAE (Direct) |
|---|---|---|---|---|---|
| Baseline | Vanilla DeepLabV3+ | Single Task | $L_{seg}$ + $L_{cons}$ | 0.717 | N/A |
| Exp A | AttnRegDeepLab | Single Task | $L_{seg}$ + $L_{cons}$ | **0.729** | N/A |
| Exp B1 | Pure Regression | Single Task | $L_{pre}$ | N/A | 0.057 |
| Exp B2 | Pure Regression | Single Task | $L_{reg}$ ($L_{pre}$+ $L_{range}$) | N/A | 0.051 |
| Exp C | AttnRegDeepLab | Full MTL | $L_{total}$ | 0.716 | **0.046** |
| Exp D | AttnRegDeepLab | Decoupled | Phase 1: $L_{seg}$ Phase 2: $L_{reg}$ | **0.729** | 0.049 |
| Exp E | AttnRegDeepLab (No-Inject) | Full MTL | $L_{total}$ | 0.678 | 0.053 |

*2) Quantitative Analysis.*

As listed in Table 1, the experimental results validate our core hypotheses regarding feature interaction and gradient dynamics.

**Validation of Feature Injection Mechanism** (Exp C vs. Exp E): To verify whether Feature Injection is the catalyst for multi-task success, we compared Exp C (With Injection) against Exp E (No Injection/Disconnected). The results indicate that under the exact same full training strategy, removing the injection mechanism caused significant performance degradation. Specifically, the segmentation Dice dropped from 0.716 to 0.678, and grading MAE worsened from 0.046 to 0.053, performing even worse than the single-task baseline. This demonstrates that merely summing the segmentation and regression losses induces severe gradient conflict. Only through the feature injection mechanism, which physically guides the segmentation decoder with global semantic features extracted by the regression head, can the optimization directions of these two tasks be coordinated, transforming conflict into mutual assistance.

**Efficacy of Weak Labels** (Exp B1 vs. Exp B2): Before evaluating the multi-task framework, we verified the contribution of the proposed Range Loss using the regression baselines. As shown in Table 1, training solely on limited paired data (Exp B1, N=*318*) resulted in a higher grading error (MAE = 0.057). However, by incorporating the large-scale grade-only dataset via interval constraints (Exp B2, N=1549), the error significantly decreased to 0.051. This improvement confirms that the Range Loss successfully extracts valuable global priors from discrete clinical grades, mitigating the data scarcity problem and justifying its application in the subsequent multi-task experiments (Exp C/D).

**Gradient Trade-off and Optimal Grading:** The results of Exp C revealed a significant shape-area tradeoff in deep learning optimization. To accommodate the regression task's strict requirement for total quantity accuracy, the backbone network sacrificed high-frequency boundary details, resulting in a slight decrease in Dice from the S1 baseline (0.729) to 0.716. However, the cost of this global optimization yielded the best precision for the regression head (MAE=0.046), outperforming the pure regression baseline (0.051) and matching the accuracy of segmentation-derived grading. This is qualitatively corroborated in Fig. 5; while the S1 Model (Exp A) produces sharper boundaries around small fragments (Column 3), the Full MTL Model (Exp C) generates slightly smoother masks (Column 4). Nevertheless, Exp C's global area estimation aligns more consistently with the ground truth grading. This indicates that if the primary clinical goal is precise numerical grading, the end-to-end automated pipeline with feature injection is the optimal choice.

**Robustness of the Decoupled Strategy:** In terms of addressing the high clinical standards for both explainability and model stability, Exp D (Frozen Finetuning) offers the most balanced solution. Although its MAE (0.049) is marginally higher than that of Exp C, it effectively maintains the optimal segmentation accuracy of Exp A (Dice 0.729), effectively insulating the visual features from regression gradient interference. This proves that the spatial features learned in Phase 1 possess strong generalization capabilities and can be directly utilized by the regression head for positive transfer, making Exp D the most robust candidate for clinical deployment.

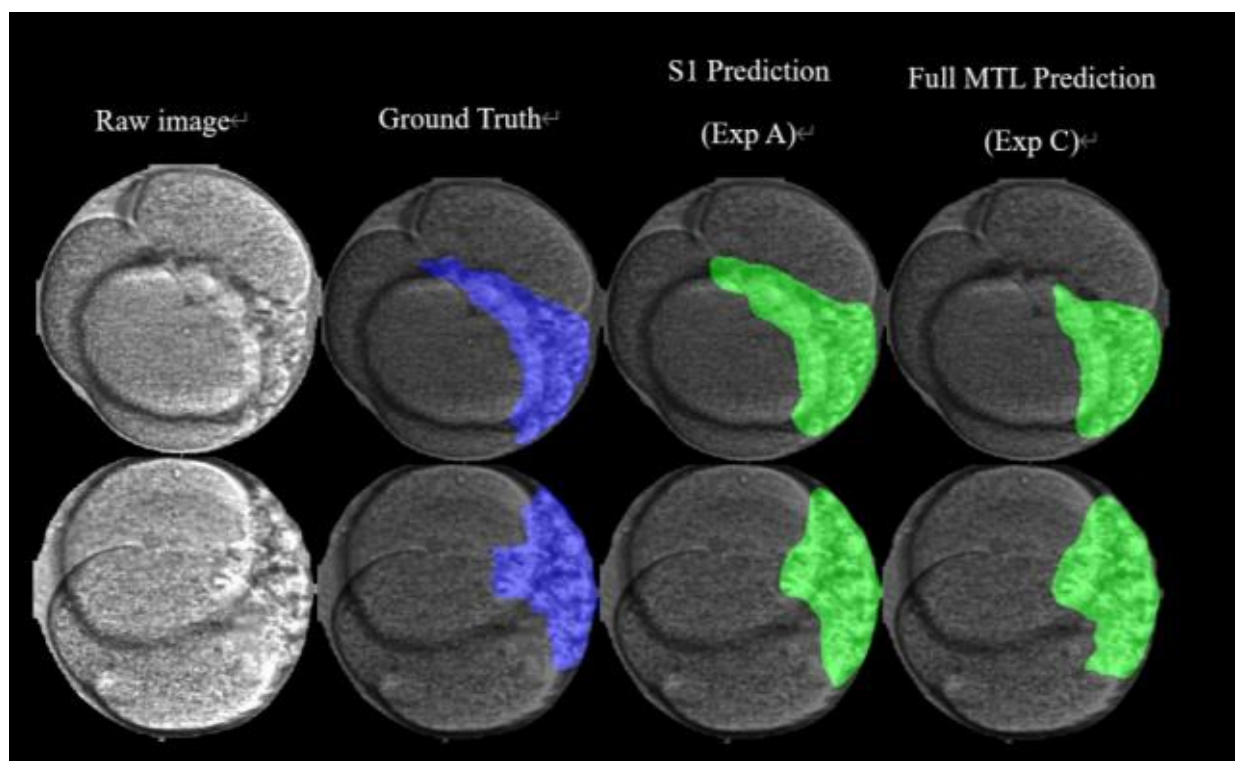

Fig. 5. Comparison of segmentation results. From left to right: (a) Raw input image. (b) Ground Truth (Blue). (c) Prediction from S1 Model (Exp A) showing sharp boundaries (Green). (d) Prediction from Full MTL Model (Exp C) showing slight over-smoothing but consistent area estimation.

Synthesizing the above findings, this study confirms that it is difficult for a single model to optimize pixel-level accuracy (DICE coefficient) and visual conformity (MAE). Therefore, we propose a Task-Specific Dual-Module Deployment Strategy deployment strategy as the optimal clinical solution: utilizing the S1 Model (Exp A) to provide the highest fidelity visual aid (XAI), while simultaneously using the Full MTL Model (Exp C) to provide the lowest error automated grading. This solution not only surpasses traditional methods and single-task baselines in technical metrics but also offers a flexible and trustworthy choice for clinical utility..

## V. Conclusion

AttnRegDeepLab successfully addresses the subjectivity and interpretability gaps in embryo fragmentation grading by recasting the task from a traditional black-box regression into a continuous quantification problem with pixel-level explainability. Driven by the failure of hand-crafted features (GLCM/LBP) on this task, we designed a semantic segmentation paradigm featuring a Feature Injection mechanism and a Two-Stage Decoupled training strategy to resolve multi-task learning (MTL) gradient conflicts. Our key findings demonstrate:

1. Architecture Necessity (Exp. A): Integrating Attention Gates and U-Net skip connections filtered cytoplasmic noise, boosting the baseline DeepLabV3+ Dice score from 0.717 to a SOTA 0.729.

2. Feature Injection Efficacy (Exp. C vs. E): Eliminating feature injection caused grading MAE to degrade from 0.046 to 0.053, proving that physical feature concatenation—not just joint loss functions—is vital for positive task transfer.

3. Gradient Trade-off (Exp. C): Minimizing grading error (MAE = 0.046) slightly compromised edge details (Dice = 0.716), confirming that pixel-level and image-level targets compete within a single architecture.

Practically, our framework serves two distinct clinical roles: a Visual Assistant setting (Dice = 0.729) for rapid anomaly localization, and an Automated Quantification setting (MAE = 0.046) to eliminate inter-observer variability. To bypass the discovered MTL gradient trade-off, future work will pivot toward a single-task learning (STL) approach focused solely on boundary preservation. This will allow us to derive fragmentation grades via an explicit, deterministic physical area formula, eliminating regression heads and gradient interference entirely to maximize clinical transparency. Finally, while the two-stage deep learning offers high accuracy, their latency is prohibitive for edge devices. Thus, the one-stage object detector achieves an optimal equilibrium between speed and precision, delivering the rapid processing required for high-throughput applications. This critical need to balance multi-stream feature fusion with strict edge-computing hardware constraints is a prominent challenge across diverse domains, mirroring recent edge-AI methodologies that deploy end-to-end automated deep fusion pipelines and localized landmark visualization [15] on restricted hardware.